\documentclass[lettersize,journal]{IEEEtran}
\usepackage{amsmath,amsfonts}
\usepackage{algorithmic}
\usepackage{algorithm}
 \usepackage[font=footnotesize]{caption}
\usepackage{booktabs}
\usepackage{array}
\usepackage[caption=false,font=normalsize,labelfont=sf,textfont=sf]{subfig}
\usepackage{textcomp}
\usepackage{stfloats}
\usepackage{url}
\usepackage{verbatim}
\usepackage{graphicx}
\usepackage{multirow}
\usepackage{siunitx}
\usepackage{makecell}
\usepackage{colortbl}
\usepackage{xcolor}
\usepackage{multicol}
\usepackage{adjustbox}
\usepackage[font=small]{caption}
\usepackage[bookmarks=true]{hyperref}
\usepackage{cite}
\begin{document}

\title{Task Editing for Generalizable 3D Visuomotor Policy Learning}

\author{
    \textbf{Jian-Jian Jiang}\textsuperscript{1,*}, 
    \quad \textbf{YiHan Yang}\textsuperscript{1,*},
    \quad \textbf{Lan Wei}\textsuperscript{1,*}, 
    \quad \textbf{Yuming Luo}\textsuperscript{4}, 
    \quad \textbf{Xiao-Ming Wu}\textsuperscript{3} \\
    \quad \textbf{Xuhang Chen}\textsuperscript{1},
    \quad \textbf{Bin Fan}\textsuperscript{1},
    \quad \textbf{Dandan Zhang}\textsuperscript{2},
    \quad \textbf{Wei-Shi Zheng}\textsuperscript{1,$\dagger$} \\
    \textsuperscript{1} \small Sun Yat-sen University
    \textsuperscript{2} \small Imperial College London 
    \textsuperscript{3} \small Nanyang Technological University 
    \textsuperscript{4} \small South China University of Technology \\
    * Equal contribution. $\dagger$ Corresponding author.
}

\markboth{Journal of \LaTeX\ Class Files,~Vol.~14, No.~8, August~2021}%
{Shell \MakeLowercase{\textit{et al.}}: A Sample Article Using IEEEtran.cls for IEEE Journals}


\twocolumn[{
\renewcommand\twocolumn[1][]{#1}
\vspace{-4mm}
\maketitle
\begin{center}
    \vspace{-10mm}
    \captionsetup{type=figure}
    \includegraphics[width=0.98\textwidth]{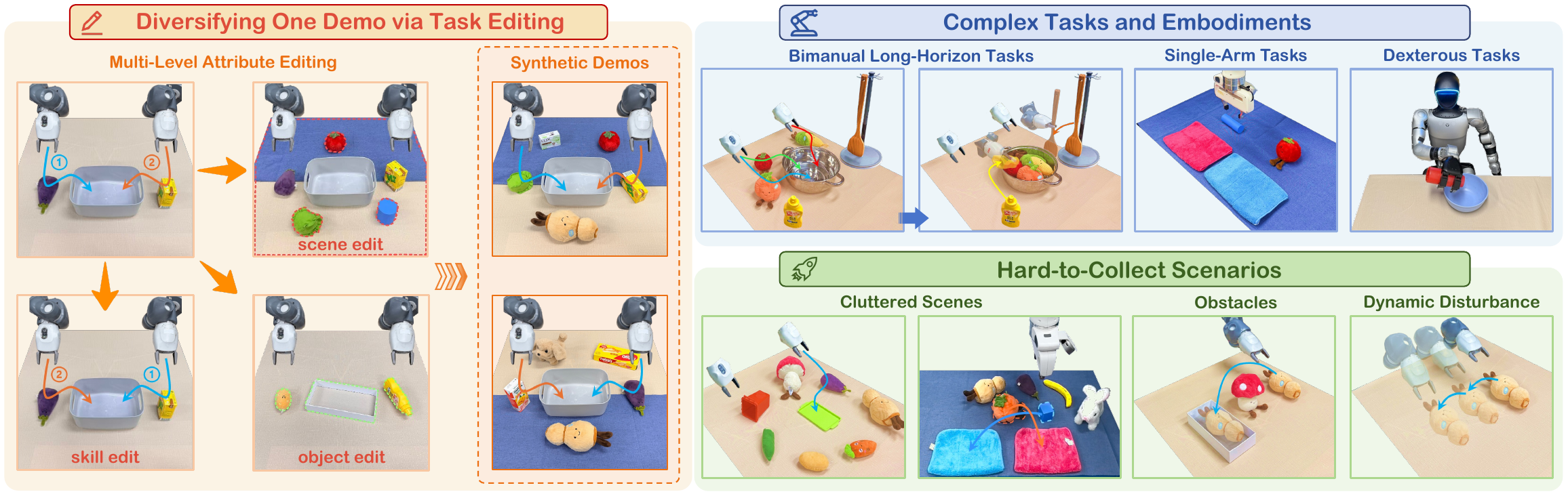}
    \caption{\textbf{Task-Edit} is a data-efficient framework that uses only a single demonstration to generate trajectories from a task-centric editing perspective. The key insight of Task-Edit lies in the decomposition of tasks into three core components: scenes, manipulation skills and objects. By editing the attributes of each component separately, Task-Edit generates diverse demonstrations for training. Furthermore, Task-Edit can be applied to various manipulation tasks and embodiments while extending easily to scenarios where data collection is difficult.}
    \label{fig:teaser}
    \vspace{-0.5mm}
\end{center}
}]

\begin{abstract}
3D visuomotor policies offer a promising direction for complex robotic manipulation, as depth maps and point clouds provide rich geometric information for spatial reasoning. However, their success often depends on large-scale real-world demonstrations, which are costly and time-consuming to collect.
To this end, existing methods commonly use demonstration generation strategies to improve data efficiency by applying object-centric transformations to human-collected demonstrations, such as varying object poses or scales. While effective for local variation, these transformations largely preserve the original scene structure and skill sequence, limiting their ability to synthesize diverse scene-skill-object combinations for complex tasks.
In this paper, we propose Task-Edit, a novel demonstration generation framework that generates diverse trajectories from a task-centric editing perspective. The key insight of Task-Edit is to decompose a task into scene, skill and object components, and flexibly recombine them. In this way, Task-Edit enables scalable demonstration generation and significantly improves generalization for long-horizon manipulation tasks.
We evaluate Task-Edit through extensive real-world experiments and demonstrate three advantages: (1) Effectiveness: Task-Edit significantly improves 3D visuomotor policies across various real-world tasks and robot embodiments. (2) Generalizability: Task-Edit improves model generalization across different scenario setups. (3) Applicability: Task-Edit enables models to handle scenarios that are difficult to collect in the real world, including disturbance resistance, obstacle avoidance and unseen cluttered scenes.
\end{abstract}


\section{Introduction}
\IEEEPARstart{R}{ecently}, 3D visuomotor policies demonstrate significant potential in complex robotic manipulation tasks \cite{medical_surgery, industrial_assembly} by using depth maps or point clouds as input for spatial perception. However, collecting such 3D data in the real world is often labor-intensive and time-consuming.

To alleviate this data bottleneck, recent demonstration generation methods \cite{mimicgen, dexmimicgen, r2r2r, cpgen, skillmimicgen, demogen, AutoTrialGen} make great progress in improving data efficiency by increasing the number of available demonstrations. These methods augment data from an object-centric perspective by varying initial object positions \cite{mimicgen, dexmimicgen, skillmimicgen, r2r2r, demogen, AutoTrialGen} or resizing objects \cite{cpgen}, achieving substantial performance improvements. However, prior studies \cite{data_scaling_laws} suggest that simply increasing demonstration quantity through object-centric transformations is insufficient for achieving policy generalization. Therefore, although previous works improve the quality and efficiency of demonstration generation, they are limited in enhancing generalization due to fixed behavioral patterns.

To address this limitation, we propose Task-Edit, a data-efficient framework for generating diverse trajectories from a task-centric perspective. Compared with previous methods, Task-Edit decouples a task into three core components: scene, skill and object. Concretely, the scene component defines the task context, including object composition and environmental layout. The skill component characterizes the manipulation primitives involved in the task, such as whether they require collaboration, whether they are cyclic and how they are sequenced in long-horizon execution. Moreover, the object component captures properties including 6-DoF pose, texture, geometry and scale, as in previous methods. The key insight of Task-Edit is that data diversity can be increased by decoupling demonstrations into scene, skill and object attributes, and editing each attribute independently. This enables synthesis of diverse scene-skill-object combinations, which is especially important for improving generalization in long-horizon manipulation tasks.
For example, in a cooking task, given one human demonstration, Task-Edit enables independent and systematic variation of the cooking setup, such as the composition and sequence of manipulation skills and the target objects or ingredients, which is not achievable with existing data generation frameworks. As a result, Task-Edit can generate diverse training trajectories from a single human demonstration, enabling more robust and generalizable 3D visuomotor policies.

To achieve this, we incorporate a Real2Sim2Real paradigm \cite{r2r2r} to edit tasks and collect data in simulation \cite{isaaclab}, providing a more flexible and efficient alternative. Unlike previous Real2Sim2Real paradigms, our pipeline precisely handles and combines the three attributes in Real2Sim and accurately recovers the scene in Sim2Real.
Specifically, in Real2Sim, we use 3D generation models \cite{rodin, hunyuan3d} to obtain diverse 3D object assets. Then, we use keypoint matching to align corresponding manipulation parts across objects and transfer skills to new objects (e.g., from tomatoes to eggplants), creating new skill-object combinations. Moreover, by modifying skill execution sequences, e.g., from ``pouring first then stirring'' to ``stirring first then pouring'', as well as 6-DoF poses and combinations of source and target objects, we generate more skill-object configurations. Finally, by varying the background and constituent objects (adding distractors), we construct new scenes and create numerous scene-skill-object combinations to generate diverse data from a single demonstration.

By utilizing simulation, we can flexibly modify tasks and efficiently collect data, including end-effector poses, joint angles, depth maps and point clouds from arbitrary viewpoints. However, directly using such data for policy training suffers from the sim-to-real gap. To address this, we introduce a sim-to-real repairing module in Sim2Real. We observe that the gap mainly arises from missing depth values and noisy real-world point clouds, which primarily occur at object edges. Based on this observation, we employ a depth-repairing foundation model \cite{depth_repair} that takes RGB and depth images as input to obtain refined depth maps. Meanwhile, we utilize a segmentation foundation model \cite{sam2} to acquire object masks for real-time object point-cloud repair. We also deploy a filtering algorithm to reduce noisy points around the end-effector and filter out background point clouds. By combining repaired object and end-effector point clouds, we obtain higher-quality point clouds, reducing the sim-to-real gap and improving downstream policy performance.

Extensive real-world experiments demonstrate the following three properties of our method: (1) Effectiveness: Task-Edit significantly enhances policy performance across various real-world manipulation tasks, especially for long-horizon tasks. Our approach is applicable to various embodiments, including the dual-arm robots ABB YuMi and Cobot Magic, the humanoid robot Unitree G1, and the single-arm robot Franka Panda. (2) Generalizability: Task-Edit improves the generalization capability of models across various scene setups involving unseen objects or novel source-target object combinations. (3) Applicability: Task-Edit enhances model applicability even in scenes that are difficult to collect in the real world, such as those requiring disturbance resistance and obstacle avoidance, or cluttered scenes containing various distractors.

\section{Related Work}
\subsection{3D Visuomotor Policy Learning}
Visuomotor policy learning provides an efficient way for robots to acquire human-like skills, as it relies on observation-action pairs from expert demonstrations. Given the challenges of estimating object states in the real world, visual observations such as images or point clouds serve as practical alternatives. While 2D image-based policies \cite{diffusion_policy, uva, pi0, openvla} predominate, the significance of 3D visuomotor policies \cite{peract, gnfactor, manigaussian, dif} is increasingly recognized. Specifically, PerAct \cite{peract} and GNFactor \cite{gnfactor} reconstruct voxel grids from multi-view RGB-D images, and fuse voxel, language and robot-state features through a Transformer to predict actions based on regression. In contrast, 3D Diffusion Policy (DP3) \cite{3d_diffusion_policy} leverages 3D point clouds as input and extracts geometric features via PointNet \cite{pointnet}, which are combined with robot proprioception to predict future actions based on a generative paradigm \cite{ddim}. While these methods possess generalization, they require many human-collected trajectories for training, incurring significant labor and time costs. To address this limitation, instead of scaling up via human labor, Task-Edit synthesizes diverse data by editing task components, reducing human effort.

\subsection{Object-Centric Data Generation for Manipulation}
Demonstration generation is a promising way to improve visuomotor policies for real-world manipulation with less human effort. One line of work, represented by MimicGen \cite{mimicgen} and its extensions \cite{dexmimicgen, skillmimicgen}, adapts human-collected demonstrations to new object-centric configurations by synthesizing execution plans. However, these methods require costly on-robot rollouts to collect augmented training data, making the process complex and inefficient. Another line of work explores simulator-free generation \cite{r2e2r, demogen, r2rgen, robosplat}. DemoGen \cite{demogen} and R2RGen \cite{r2rgen} directly edit 3D point clouds to synthesize visual observations, but are limited by noise and missing depth in real-world point clouds. RoboSplat \cite{robosplat} and Real2Edit2Real \cite{r2e2r} use 3D Gaussian Splatting or generative models for rendering, but require either dense viewpoints for reconstruction or high computational cost for generation.

With great progress in simulators \cite{isaaclab}, the Real2Sim2Real paradigm enables efficient demonstration generation \cite{r2r2r, twinaligner}. Real2Render2Real \cite{r2r2r} ignores dynamics and directly sets object and robot poses for fast rendering, while TwinAligner \cite{twinaligner} aligns robot states and point clouds to reduce the sim-to-real gap. However, they primarily focus on object-centric augmentation, such as changing initial object poses or scales. This limits data diversity and leaves the potential of simulation underexplored.
To address this issue, we propose Task-Edit, a simulator-based Real2Sim2Real framework that generates diverse training data by decomposing and reconfiguring tasks. Task-Edit decouples each task into scenes, manipulation skills and objects, and edits them separately in simulation to fully exploit the advantages of simulators.

\begin{figure*}[t]
  \centering
  \includegraphics[width=0.98\textwidth]{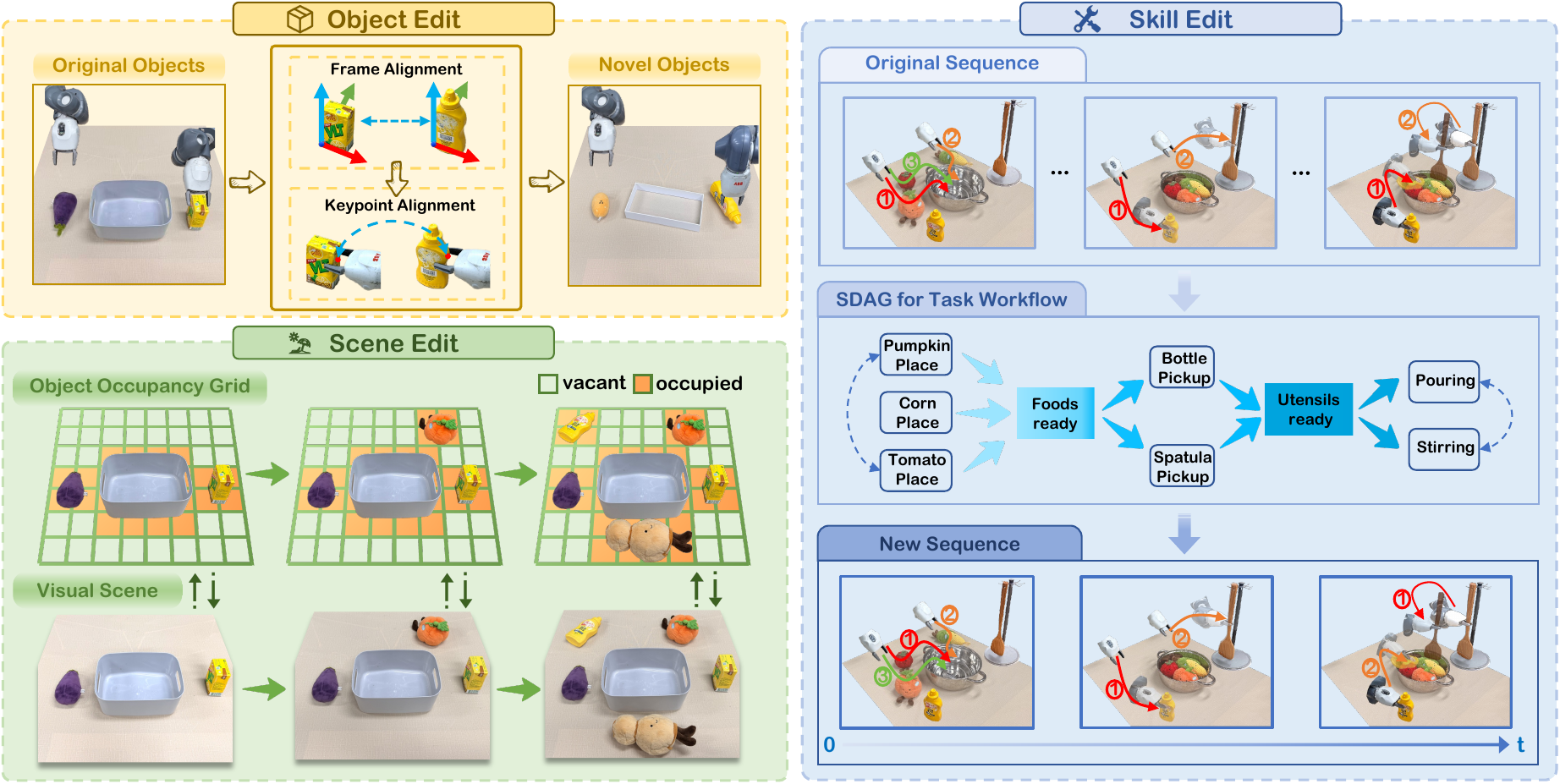}
  \caption{\textbf{Architecture of the Task-Edit Framework.} Our framework decomposes tasks into scenes, manipulation skills and objects, and edits each component independently. For object editing, Task-Edit aligns object origins and uses keypoint matching with spatial transformations to transfer skills to new objects. For skill editing, it converts the source demonstration into a Skill Directed Acyclic Graph and identifies interchangeable segments based on DAG properties and robot constraints. For scene editing, Task-Edit adds distractor objects with a divide-and-conquer strategy that partitions the workspace and manages space use through occupancy maps.}
  \label{fig:pipeline}
  \vspace{-3mm}
\end{figure*}

\section{Task-Edit Approach}
\subsection{Problem Formulation of Task-Edit}

A 3D visuomotor policy $\pi: \mathcal{O} \mapsto \mathcal{A}$, learned from a demonstration dataset $\mathcal{D}$, takes observations $o \in \mathcal{O}$, such as point clouds or depth maps, and proprioception as inputs to predict actions $a \in \mathcal{A}$. We define a source human-collected demonstration conditioned on an initial scene-skill-object configuration $c_0 \in \mathcal{C}$ as $D_{c_0} = (d_0^0, d_1^0, \dots, d_{L-1}^0 | c_0)$, where each $d_t^n = (o_t^n, a_t^n)$ represents an observation-action pair. Here, $n$ and $L$ denote the configuration index and length respectively. Task-Edit edits this demonstration by modifying the attributes of the original scene $e_0$, manipulation skills $m_0$ and objects $i_0$ within configuration $c_0$ to create a novel configuration $c_1$: 
\begin{equation}
    D_{c_1}=(d_0^1, d_1^1, \dots , d_{L-1}^1 | c_1).
\end{equation}

Specifically, we assume the source demonstration of a task involves $K$ objects with initial 6-DoF poses ${T_0^1, T_0^2, \dots, T_0^K}$. The action $a_t$ has two representations: the SE(3) end-effector (EE) pose in the world frame and joint angles, where the latter are computed via inverse kinematics from EE poses. The observation $o_t$ consists of vision-based data, such as point clouds and depth images, and proprioception, including EE poses and joint angles. We define the observation as $o_t = (o_t^{vision}, o_t^{prio})$, where $o_t^{prio}$ reflects the current robot state and has the same dimension as the corresponding actions.

\subsection{The Real2Sim Process}

\noindent \textbf{Reconstruct Motion and Skill Segments.}
With recent advances in image-based 3D generation \cite{rodin, hunyuan3d}, we efficiently generate 3D assets using four multi-view images of each object. Given a source human-collected demonstration, we first use LangSAM \cite{lang-segment-anything} to segment objects in the initial frame. We then employ FoundationPose \cite{foundationpose}, together with the object masks and 3D models, to estimate their 6-DoF trajectories throughout the sequence. Since we align the world coordinate frame with the robot base frame, the EE pose trajectories obtained via the robot interface directly correspond to robot motion trajectories in the simulation world frame. By integrating the reconstructed object and robot trajectories, we obtain a faithful simulation replay of the original demonstration, which serves as the basis for subsequent editing.

Following MimicGen and related work \cite{mimicgen, skillmimicgen}, we decompose each trajectory into an ordered sequence of motion and skill segments. Motion segments capture approach and transit phases, such as moving the end-effector toward an object. In contrast, skill segments correspond to on-object interactions, including grasping, stirring and shaking.

\noindent \textbf{Object Attributes Editing.}
To create novel scene-skill-object configurations, we first focus on object attribute editing, which involves modifying 6-DoF object poses, textures and geometries. Adapting actions to novel 6-DoF object poses is mainly achieved through the relative transformation between the target and source poses. This transformation is represented by a set of $4 \times 4$ homogeneous matrices $\Delta_T$ calculated as:
\begin{equation}
    \Delta_T = {(T_0^1)^{-1} \cdot T_1^1, \dots, (T_0^k)^{-1} \cdot T_1^k},
\end{equation}
Since the end-effector must maintain a consistent relative pose with the object to replay skill segments, both motion and skill segments undergo the spatial transformations based on the new object poses to generate updated robot trajectories. Taking the starting point of a skill segment $A^{EE}[\tau_0^s]$ as a reference, its transformed coordinates are expressed as:
\begin{equation}
    A^{EE}[\tau_1^s] = A^{EE}[\tau_0^s] \cdot \Delta_{T_1}.
\end{equation}

For motion segments, we utilize linear and Slerp interpolation to synthesize a smooth trajectory from the current end-effector position to $A^{EE}[\tau_1^s]$. For skill segments, we apply the spatial transformations $\Delta_{T_1}$ to the original trajectory segments to obtain the final on-object manipulation sequence.

Following that, to adapt actions to objects with diverse textures and geometries, we utilize keypoint matching to identify corresponding functional parts across different instances to facilitate skill transfer. First, we align the origins of different objects within a unified coordinate frame. We then employ keypoint matching tools to identify manipulation keypoints across these objects. Specifically, for geometrically similar instances such as various drink boxes or cups, we utilize existing matching algorithms \cite{DenseMatcher} to establish correspondences. For objects that exhibit significant geometric variations, we manually define and align keypoints to ensure precise skill transfer. Assuming that the transformation matrix between the 6-DoF poses of the source keypoint $k_0$ and the target keypoint $k_1$ is defined relative to the aligned object origin, the transformation $\Delta T_{k_1}$ is calculated as:
\begin{equation}
    \Delta T_{k_1} = (T_{0}^{k_0})^{-1} \cdot T_{0}^{k_1},
\end{equation}
Therefore, the starting point of the skill segments $A^{EE}[\tau_0^s]$ on the new object is determined by combining the global object transformation with the local keypoint shift as follows:
\begin{equation}
    \hat{A}^{EE}[\tau_1^s] = A^{EE}[\tau_0^s] \cdot \Delta_{T_1} \cdot \Delta T_{k_1}.
\end{equation}

By modifying motion and skill segments based on the 6-DoF poses, textures and geometries of objects, we synthesize novel skill-object configurations and generate trajectories.

\noindent\textbf{Skill Attributes Editing.}
To further generate novel configurations, we consider skill attribute editing, which mainly focuses on modifying skill sequences. To this end, Task-Edit represents these sequences as a Skill Directed Acyclic Graph (SDAG) to identify valid permutations of segments that can be reordered without violating task constraints.

Specifically, we define skills within a configuration $m_0$ as $\{m_0^1, m_0^2, \dots, m_0^K\}$, modeled by an SDAG $G = (M, E)$. An edge $e_{i,j} \in E$ indicates a precedence dependency between skill segments $m_0^i$ and $m_0^j$. In particular, for dual-arm synchronized skills such as lifting, a segment is triggered only after all preceding single-arm skills reach a completed state. To construct this SDAG, the skill nodes $M$ are determined during demonstration collection, while the edges $E$ and the reordered skill sequences are generated by GPT 5.5. Leveraging the properties of the SDAG, we identify skills that can be reordered based on the following rules:

(1) Two segments $m_0^i$ and $m_0^j$ are candidates for reordering if no directed path exists between them in the SDAG.

(2) For segments involving different robotic arms, reordering is permissible only if the new sequence remains collision-free and avoids spatial interference between the two arms.

By combining object and skill editing, we synthesize task trajectories that encompass both the manipulation of novel objects and varied execution sequences. This approach effectively enhances the diversity of the generated trajectories and provides a richer dataset for 3D visuomotor learning.

\noindent\textbf{Scene Attributes Editing.}
Furthermore, we effectively enhance data diversity by replacing scene backgrounds and modifying the composition of objects within the workspace. These scene-centric augmentations improve the robustness of the model across various environmental settings.

Specifically, Task-Edit discretizes the tabletop area $\mathcal{A}$ into a 2D occupancy grid $\mathcal{M}$, where each cell represents a basic unit of physical space. Initially, task-relevant object assets are instantiated into the environment to form a base occupancy map. We then randomly select several distractor objects from the asset library to fill the scene based on a divide-and-conquer allocation principle, which partitions the tabletop into several sub-regions, each maintaining its own local occupancy map. When an object is placed, its 3D bounding box is projected onto the 2D plane to update the corresponding local occupancy map. To further prevent physical collisions, we introduce a buffer distance $\delta$ such that no other objects can be placed within the exclusion area $\mathcal{E}$ defined as:
\begin{equation}
    \mathcal{E} = \{p \in \mathcal{A} \mid \text{dist}(p, \mathcal{C}_{occupied}) < \delta\}.
\end{equation}

If a feasible placement that satisfies these occupancy constraints is identified within a sub-region, the object is positioned at that location and the global occupancy map is updated. Furthermore, as the origins of the coordinate systems for different objects are pre-aligned, placing these distractor objects only requires applying translation offsets within the $xy$-plane to integrate them into the scene.

\subsection{The Sim2Real Process}

Single-view point clouds acquired in simulation are accurate and free of missing depth information, creating a discrepancy from those obtained in the real world \cite{point_cloud_completion}. Although prior grasp detection works \cite{ASGrasp} show that extensive simulation data can partially bridge the sim-to-real gap, they are insufficient for robotic manipulation due to challenges such as dynamic occlusions between robots and objects. We observe that for common diffuse-reflective objects, point cloud incompleteness and noise occur around object boundaries. Based on this observation, we develop a point cloud restoration scheme to improve the robustness of policies trained with simulated data.

Specifically, we align the world coordinate system with the robot base to define the tabletop workspace. We then employ LangSAM \cite{lang-segment-anything} to segment tabletop objects and obtain their masks. Following advances in depth completion, we leverage \cite{depth_repair} to restore depth maps and use object masks to extract restored object point clouds. Moreover, we use SAM2 \cite{sam2}, a video segmentation foundation model, for real-time object mask acquisition to repair object point clouds. For the remaining point clouds, we apply a Statistical Outlier Removal (SOR) filter to eliminate flying noise around the robotic arm and use DBSCAN clustering \cite{dbscan} to remove residual tabletop point cloud clusters. This process produces high-quality real-world point clouds and reduces the sim-to-real gap.

\subsection{Overall Framework of Task-Edit}
\noindent\textbf{Overall Framework.}
Bringing all together, we develop the Task-Edit framework. We leverage FoundationPose \cite{foundationpose} to extract 6-DoF object pose trajectories, enabling reconstruction of human-collected demonstrations in simulation. During the Real2Sim process, we achieve object-level skill transfer to various instances through keypoint matching. At the skill level, we perform temporal editing of long-horizon tasks based on Skill Directed Acyclic Graphs. At the scene level, we generate cluttered scenarios using the divide-and-conquer allocation principle. Together, these components enable generation of diverse configurations and enriched task trajectories. To narrow the sim-to-real gap during Sim2Real, we leverage depth completion models and filtering algorithms to restore scene point clouds. This addresses point cloud noise and missing depth values, enhancing the performance of 3D policies.

\noindent\textbf{Implementation Details.}
We partition the table into three sub-regions, where each unit measures $2 \times 2$ cm and the buffer zone $\delta$ is 4 cm. We train for 2000 epochs using four NVIDIA RTX 4090 GPUs and AdamW with a batch size of 64.

\begin{table*}[t]
    \centering
    \caption{\textbf{Performance comparison across nine real-world tasks under in-distribution settings.} Best results are highlighted in bold.}
    \label{tab:real_world}

    \begin{adjustbox}{width=\textwidth,center}
        \begin{tabular}{l cccc ccc cc c}
            \toprule
            \multirow{2}{*}{\textbf{Method}} & \multicolumn{4}{c}{\textbf{ABB}} & \multicolumn{3}{c}{\textbf{Franka}} & \multicolumn{2}{c}{\textbf{Humanoid}} & \multirow{2}{*}{Average} \\ 
            \cmidrule(lr){2-5} \cmidrule(lr){6-8} \cmidrule(lr){9-10}
             & Barbecue & Cooking & Lifting & Sorting & Table Cleaning & Sorting & Insertion & Large Object Grasping & Pouring & \\ 
            \midrule

            DemoGen \cite{demogen} & 0.600 & 0.600 & 0.750 & 0.667 & 0.600 & 0.750 & 0.600 & 0.733 & 0.600 & 0.656 \\

            Human-Collected & 0.650 & 0.650 & 0.708 & 0.733 & 0.750 & 0.750 & 0.733 & 0.800 & 0.700 & 0.719 \\ 

            \midrule

            Ours & \textbf{0.700} & \textbf{0.750} & \textbf{0.833} & \textbf{0.867} & \textbf{0.800} & \textbf{0.900} & \textbf{0.800} & \textbf{0.867} & \textbf{0.800} & \textbf{0.813} \\ 
            \bottomrule
        \end{tabular}
    \end{adjustbox}
    \vspace{-3mm}
\end{table*}

\section{Experiments}
\subsection{Experimental Setup}
\noindent\textbf{Hardware.}
In this paper, we use four robots: the dual-arm robots ABB YuMi and AgileX Cobot Magic, the humanoid robot Unitree G1, and the single-arm robot Franka Panda. The humanoid robot uses dexterous hands, while the others use parallel-jaw grippers. For real-world experiments on YuMi, G1 and Panda, we utilize a RealSense L515 depth camera in a third-person view to collect point cloud observations. For simulation experiments on Cobot Magic, we use a RealSense D435 camera to capture third-person observations.

\noindent\textbf{Dual-Arm Tasks.}
We leverage Task-Edit to edit six dual-arm tasks: (1) \textit{Barbecue}: The left arm places vegetables on a grill while the right arm applies sauce. (2) \textit{Cooking}: Both arms place vegetables into a pot, after which the left arm pours sauce while the right arm stirs with a spatula. (3) \textit{Lifting}: The arms each place an object into a container and lift the container together. (4) \textit{Large Object Grasping}: Both arms grasp different large objects. (5) \textit{Meal Preparation}: The robot places burgers and fries onto a tray. (6) \textit{Handover}: The right arm grasps a microphone on the table and hands it to the left arm.

\noindent\textbf{Single-Arm Tasks.}
We also leverage Task-Edit to edit six single-arm tasks: (1) \textit{Sorting (ABB)}: The robot sorts objects into bins by color. (2) \textit{Table Cleaning}: The robot places tabletop objects into a container. (3) \textit{Sorting (Franka)}: The robot places objects of different colors onto corresponding towels. (4) \textit{Insertion}: The robot inserts cylinders, hexagonal prisms or cuboids into corresponding slots. (5) \textit{Pouring}: The robot grasps a bottle or goblet and pours water into a bowl. (6) \textit{Cup Placement}: The robot places different cups onto coasters.

\noindent\textbf{Baseline.}
We select DemoGen \cite{demogen}, a representative data generation method, as our main baseline. DemoGen mainly applies spatial transformations to end-effector and object point clouds to generate large-scale trajectories across various positions. We also use human-collected trajectories as another baseline. After data preparation, we use DP3 \cite{3d_diffusion_policy}, a representative 3D visuomotor policy, and train it on these distinct data sources.

\subsection{Evaluation on In-Distribution Scenarios}
\noindent\textbf{Quantitative Experiments in Real World.}
We evaluate DP3 \cite{3d_diffusion_policy} on Task-Edit data, DemoGen data \cite{demogen} and human-collected data, with results reported in Tab. \ref{tab:real_world}. Notably, Task-Edit generates trajectories for various combinations using \textbf{only a single demonstration}. In contrast, DemoGen augments data from a pre-collected demonstration set covering different combinations. Except for Lifting, which contains 120 generated trajectories, all other tasks contain 60. For the human-collected baseline, we collect 50 demonstrations per task.

As shown in Tab. \ref{tab:real_world}, DP3 achieves SOTA performance when trained on Task-Edit data, outperforming both DemoGen and human-collected data. We attribute this success to our ability to obtain higher-quality point clouds in simulation. Furthermore, our Sim2Real process effectively reduces the sim-to-real gap and facilitates DP3 deployment. In contrast, since DemoGen directly edits real-world 3D data, both DemoGen and the human-collected baseline suffer from depth noise and missing depth values inherent in depth cameras.

\begin{table*}[t]
    \centering
    \caption{\textbf{Performance comparison on three simulation tasks across in- and out-of-distribution settings.} Best results in bold.}
    \label{tab:simulation}

    \setlength{\tabcolsep}{7pt}
    
    \begin{tabular}{l cccc c ccc}
        \toprule
        \textbf{Method} & \multicolumn{4}{c}{\textbf{In-Distribution}} & \multicolumn{4}{c}{\textbf{Out-Of-Distribution}} \\
        \cmidrule(lr){2-5} \cmidrule(lr){6-9}
        & Meal Preparation & Handover & Cup Placement & Average & Meal Preparation & Handover & Cup Placement & Average \\
        \midrule

        DemoGen \cite{demogen} & 0.490 & 0.800 & 0.640 & 0.643 & 0.420 & 0.820 & 0.510 & 0.583 \\

        Human-Collected & 0.620 & 0.990 & 0.990 & 0.867 & 0.660 & 1.000 & 0.830 & 0.830 \\

        \midrule

        Ours & \textbf{0.810} & \textbf{1.000} & \textbf{1.000} & \textbf{0.937} & \textbf{0.730} & \textbf{1.000} & \textbf{0.930} & \textbf{0.887} \\
        \bottomrule
    \end{tabular}
    \vspace{-2mm}
\end{table*}

\begin{figure*}[t]
  \centering
  \includegraphics[width=0.95\textwidth]{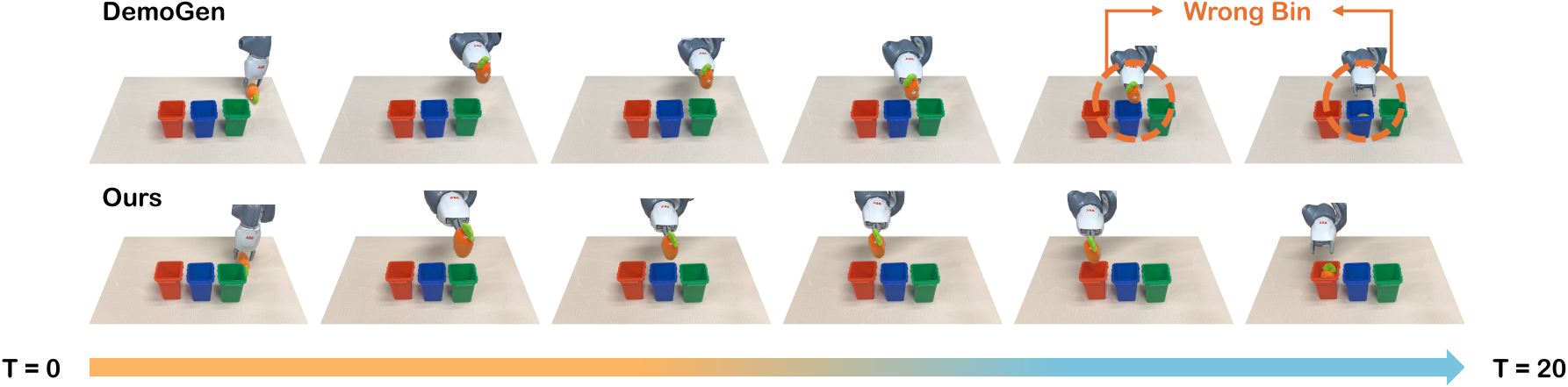}
  \caption{\textbf{Visualization of execution in the Sorting (ABB) task.} This time-series visualization compares the execution process of DP3 trained on DemoGen data \cite{demogen} and Task-Edit data in the Sorting task.}
  \label{fig:visualization1}
  \vspace{-2.5mm}
\end{figure*}

\begin{figure}[t]
  \centering
  \includegraphics[width=0.97\linewidth]{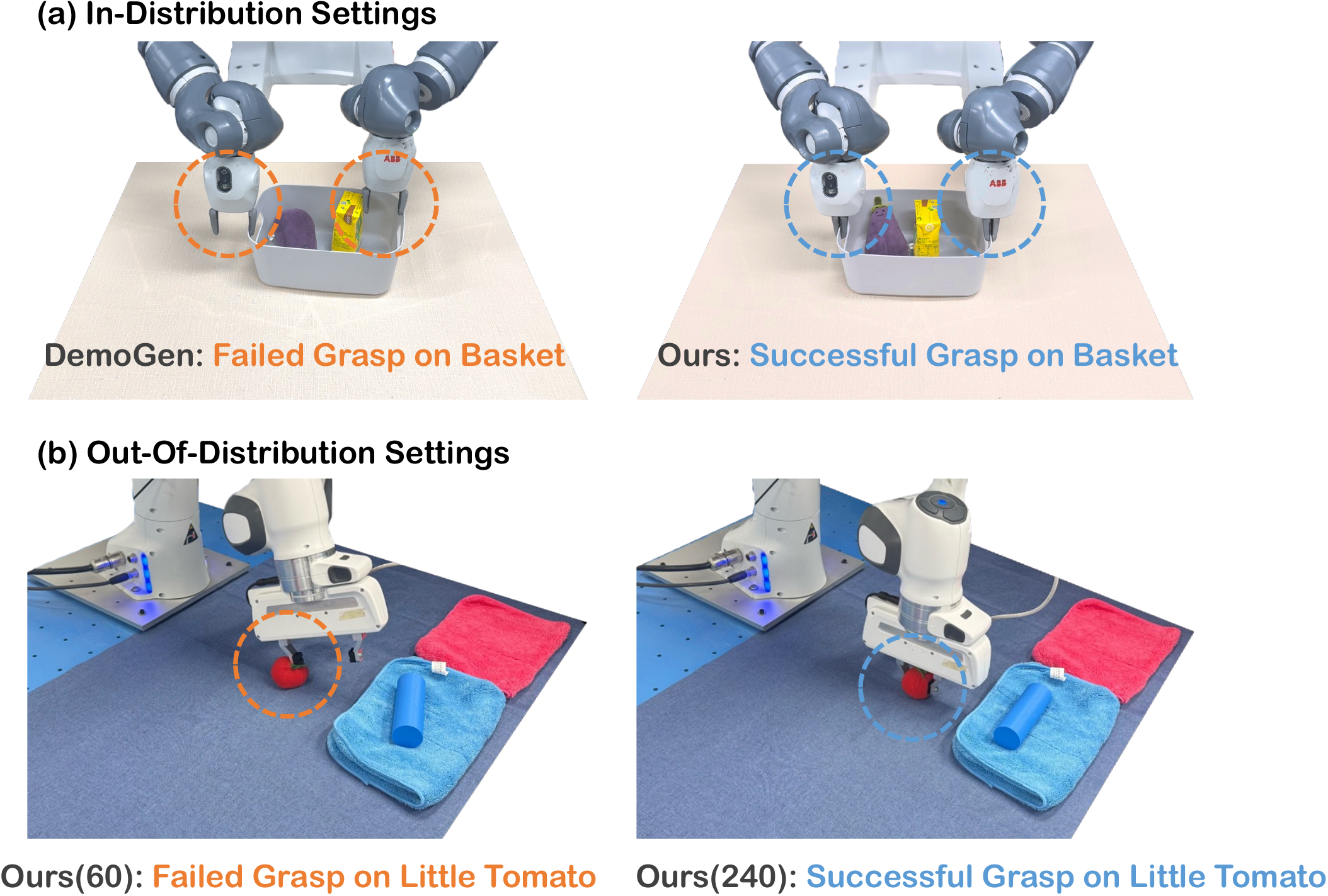}
  \caption{\textbf{Qualitative Results.} In Fig. (a), we compare DP3 models trained on DemoGen and Task-Edit data in in-distribution Lifting. The model trained with DemoGen data fails to grasp the container edge stably, while our method achieves a steady grasp. In Fig. (b), we show DP3 performance on the Sorting task under an out-of-distribution setting. As the data scale and object-combination diversity increase, the model trained on 240 trajectories generates more stable grasping motions for unseen objects such as a cherry tomato.}
  \label{fig:visualization2}
  \vspace{-4.5mm}
\end{figure}

\noindent \textbf{Qualitative Experiments in Real World.}
We visualize DP3 models trained on data generated by Task-Edit and DemoGen \cite{demogen} respectively. As illustrated in Fig. \ref{fig:visualization1} and Fig. \ref{fig:visualization2}(a), the model trained on DemoGen data incorrectly reaches the target destination and fails to complete tasks, whereas the model trained on Task-Edit data achieves better performance. We attribute this to Task-Edit obtaining higher-quality point clouds from simulation. During deployment, our restoration module effectively improves real-world point cloud quality.

\noindent\textbf{Quantitative Experiments in Simulation.}
We evaluate DP3 \cite{3d_diffusion_policy} on data generated by Task-Edit, data generated by DemoGen \cite{demogen} and human-collected data, with results reported in Tab. \ref{tab:simulation}. Except for Meal Preparation, which uses 60 trajectories, the other two tasks use 200 trajectories. The amount of human-collected data equals that of generated data.

As illustrated in Tab. \ref{tab:simulation}, although Task-Edit uses \textbf{only one source sample} for generation, it still achieves SOTA performance. We attribute this to the fact that, unlike DemoGen \cite{demogen}, which directly edits raw point clouds, Task-Edit avoids point-cloud incompleteness caused by affine transformations and single-view observations. Compared with human-collected data, Task-Edit can define the initial object position distribution, reducing the imbalance caused by random placement.

\subsection{Evaluation on Generalization Scenarios}

\noindent\textbf{Experiments in Real World.}
We evaluate the generalization capability of DP3 \cite{3d_diffusion_policy} across varying trajectory scales (60, 120, and 240) generated by Task-Edit and DemoGen \cite{demogen}, with the results reported in Tab. \ref{tab:generalization}. Notably, Task-Edit synthesizes all training data using \textbf{only a single human demonstration}, whereas DemoGen uses multiple human demonstrations. Moreover, both the objects and their combinations in this evaluation remain unseen during training. As shown in Tab. \ref{tab:generalization}, DP3 generalization improves as object and combination diversity increase with the number of generated trajectories, and Task-Edit consistently outperforms DemoGen. We also visualize models trained with different numbers of trajectories in the Sorting (Franka) task. As shown in Fig. \ref{fig:visualization2}(b), larger trajectory sets cover more diverse objects, enabling the trained models to more precisely grasp objects of different scales, such as small tomatoes unseen during training.

\noindent\textbf{Experiments in Simulation.}
We evaluate the performance of DP3 \cite{3d_diffusion_policy} on data generated by Task-Edit, data generated by DemoGen \cite{demogen} and human-collected data, with the results reported in Tab. \ref{tab:simulation}. The generation settings of Task-Edit and DemoGen are aligned with real-world processing, and the amount of human-collected data is the same as that of the generated data. The experimental conclusions are largely consistent with those in the in-distribution setting.

\begin{table}[t]
\centering
\caption{\textbf{Generalization performance comparison between Task-Edit and DemoGen across varying trajectory scales.}}
\label{tab:generalization}

\setlength{\tabcolsep}{12pt} 

\begin{tabular}{@{} l S[table-format=1.3] S[table-format=1.3] S[table-format=1.3] S[table-format=1.3] @{}}
\toprule
\multirow{2.5}{*}{\textbf{Method}} 
    & \multicolumn{2}{c}{\textbf{ABB}} 
    & {\textbf{Franka}} 
    & {\multirow{2.5}{*}{\textbf{Average}}} \\
\cmidrule(lr){2-3} \cmidrule(lr){4-4}
    & {Lifting} & {Cooking} & {Sorting} & \\
\midrule

DemoGen (60)  
    & {0.423} & 0.500 & {0.550} & {0.485} \\
Ours (60)   
    & \textbf{0.538} & \textbf{0.600} & \textbf{0.650} & \textbf{0.591} \\

\midrule 

DemoGen (120) 
    & 0.538 & 0.600 & {0.650} & {0.591} \\
Ours (120)  
    & \textbf{0.692} & \textbf{0.700} & \textbf{0.800} & \textbf{0.727} \\

\midrule

DemoGen (240) 
    & 0.654 & 0.650 & {0.700} & {0.667} \\
Ours (240)  
    & \textbf{0.846} & \textbf{0.750} & \textbf{0.850} & \textbf{0.818} \\

\bottomrule
\end{tabular}
\vspace{-3mm}
\end{table}

\subsection{Evaluation on Tasks with Edited Skill Sequences}
We evaluate DP3 \cite{3d_diffusion_policy} trained on various skill sequences generated by Task-Edit \textbf{using a single demonstration} versus those augmented by DemoGen \cite{demogen} using a demonstration set covering each sequence, with results reported in Tab. \ref{tab:skill_sequence}. Specifically, Cooking is a dual-arm long-horizon task, while Sorting is a single-arm short-horizon task. As shown in the table, DP3 trained on our data outperforms DemoGen in both task categories. We attribute this to the fact that human-collected trajectories for different sequences often contain varying preferences, causing inconsistencies that make visuomotor policies harder to learn manipulation skills. In contrast, Task-Edit uses only one human-collected demonstration for editing and augmentation. Therefore, the generated trajectories remain consistent with the source demonstration, making it easier for the policy to acquire manipulation skills.

\begin{table}[t]
\centering

\setlength{\tabcolsep}{11pt} 

\caption{\textbf{Performance comparison across two tasks under different skill-sequence settings.} The Cooking task involves six edited sequences, while the Sorting (Franka) task has four.}
\label{tab:skill_sequence}

\begin{tabular}{l S[table-format=1.3] S[table-format=1.3] S[table-format=1.3]}
\toprule
\textbf{Method} & {Cooking} & {Sorting (Franka)} & {Average} \\ 
\midrule

DemoGen \cite{demogen} & 0.667 & 0.700 & 0.684 \\ 

Ours & \textbf{0.833} & \textbf{0.950} & \textbf{0.895} \\ 

\bottomrule
\end{tabular}
\end{table}

\begin{table}[t]
\centering
\setlength{\tabcolsep}{8pt} 

\caption{\textbf{Performance of cyclic manipulation in in-distribution and generalization settings.} The source data consists of a single demonstration where the robot picks up a box and shakes it once.}
\label{tab:cycles}

\begin{tabular}{l S[table-format=1.3] S[table-format=1.3] S[table-format=1.3] S[table-format=1.3]}
\toprule
\textbf{Evaluation Setup} & {1 cycle} & {2 cycles} & {3 cycles} & {Average} \\ 
\midrule
Ours-ID & 0.933 & 0.867 & 0.800 & 0.867 \\ 
Ours-OOD & 0.733 & 0.733 & 0.667 & 0.711 \\ 
\bottomrule
\end{tabular}
\vspace{-3mm}
\end{table}

\subsection{Evaluation on Cyclic Manipulation}
Cyclic manipulation is common in daily life, such as cutting vegetables or shaking bottles. However, these tasks require demonstrations involving many different objects and various numbers of cycles, and collecting human data for all these combinations is prohibitively expensive. In this paper, we utilize Task-Edit to edit the objects and skills in cyclic manipulation tasks, which effectively reduces the human labor required for generating diverse data. In this setting, we only collect \textbf{one demonstration} of a robot shaking \textbf{a drink box once}. With Task-Edit, we generate data where the robot shakes different drink boxes for any number of cycles. To evaluate the utility of these data, we examine the success rates of DP3 \cite{3d_diffusion_policy} when shaking seen and unseen objects for 1, 2 and 3 cycles. The results are reported in Tab. \ref{tab:cycles}. The table illustrates that the visuomotor policy uses our data effectively, and these data help improve its generalization ability.

\begin{table}[t]
    \centering
    \caption{\textbf{Evaluation in hard-to-acquire scenarios.}}
    \label{tab:robustness}

    \setlength{\tabcolsep}{9pt}
    \sisetup{table-format=1.3, table-number-alignment=center}

    \begin{tabular}{@{} l *{4}{S} @{}}
        \toprule
        \textbf{Method} & 
        {Disturbance} & 
        {Distractors} & 
        {Clutter} & 
        {Average} \\
        \midrule

        Ours w/o Editing & 0.200 & 0.250 & 0.450 & 0.300 \\
        Ours    & \textbf{0.933} & \textbf{0.700} & \textbf{0.800} & \textbf{0.811} \\
        
        \bottomrule
    \end{tabular}
\end{table}

\subsection{Evaluation on Hard-to-Acquire Scenarios}
Currently, most experiments are conducted in ideal, clean scenarios. However, real-world applications involve cluttered and dynamic scenes, requiring robots to handle obstacle avoidance, moving objects, occlusions and close distractors. Due to the variability and complexity of these scenarios, relying solely on manual data collection is inefficient and lacks diversity. Inspired by DemoGen \cite{demogen}, we enable Task-Edit to generate data for these challenging settings. Moreover, extensive simulation asset libraries help generate richer and more diverse training data.

We report results for dynamic disturbances, close distractors and cluttered scenes in Tab. \ref{tab:robustness}. Since these scenarios mainly involve object movement or occlusions, models trained in ideal environments still retain a baseline ability to complete them. In contrast, obstacle avoidance requires the robot to execute lift-over motions adapted to the heights of different obstacles. Without such data in training, the model almost entirely fails. By placing obstacles into the scene and editing robot and object trajectories, Task-Edit generates diverse trajectories for navigating various obstacles and achieves a \textbf{92\% success rate}. As illustrated in Tab. \ref{tab:robustness}, using Task-Edit data significantly improves performance, demonstrating its effectiveness in generating data for hard-to-collect scenarios and enhancing performance in these settings.

\begin{table}[t]
    \centering
    \caption{\textbf{Systematic analysis of Task-Edit.} We conduct a sensitivity analysis on its trajectory reconstruction error and evaluate the impact of the depth restoration module on the overall system.}
    \label{tab:sensitivity}
    
    \setlength{\tabcolsep}{6pt}

    \begin{tabular}{lcc}
        \toprule
        \textbf{Sensitivity \& Ablation Study} & Cooking (60) & Sorting ABB (60) \\
        \midrule
        
        Ours w/ 0.003 m Random Noise & 0.700 & 0.800 \\
        Ours w/ 0.001 m Random Noise & 0.750 & 0.867 \\
        Ground Truth                 & 0.750 & 0.867 \\
        \midrule
        DemoGen \cite{demogen}       & 0.600 & 0.667 \\
        Ours w/o Depth Restoration   & 0.700 & 0.800 \\
        \midrule
        
        Ours                         & 0.750 & 0.867 \\
        
        \bottomrule
    \end{tabular}
    \vspace{-3mm}
\end{table}

\subsection{Systematic Analysis of Task-Edit}

\noindent \textbf{Sensitivity to Pose Estimation Errors.}
We use marker-based tracking to obtain ground-truth 6-DoF object poses and find that FoundationPose achieves average errors of 0.005 m and 2.236$^\circ$. By adding Gaussian noise to FoundationPose-estimated poses, we evaluate DP3 performance across different trajectories, with results for noisy FoundationPose-estimated trajectories and ground-truth trajectories reported in Tab. \ref{tab:sensitivity}.

\noindent \textbf{Ablation Study of Depth Restoration.}
As shown in Tab. \ref{tab:sensitivity}, the depth restoration module effectively mitigates the Sim-to-Real gap. Notably, even without this module, models trained on Task-Edit data still outperform those trained with DemoGen \cite{demogen}. This advantage comes from DemoGen directly applying the transformation matrix $\Delta_T$ to point clouds, which may cause missing points from the current camera view. In contrast, since Task-Edit collects data in simulation, it guarantees point-cloud completeness from any camera view, yielding higher-quality training data than DemoGen.

\noindent \textbf{Time Cost of Task-Edit.}
The overall pipeline of Task-Edit includes both manual and automated processes. Specifically, human intervention is required for collecting a single demonstration, aligning object coordinate frames and matching keypoints across objects with large geometric differences (e.g., round apples and elongated corn). When generating 300 trajectories for the long-horizon Cooking task, human intervention takes 885 s out of a total 6067.440 s, whereas collecting equivalent human demonstrations requires 90000 s. Consequently, human intervention accounts for 14.59\% of the total data generation process, yielding a 93.26\% improvement in data collection efficiency. Furthermore, the DP3 model with depth restoration reaches a control frequency of 5.26 Hz on a laptop NVIDIA RTX 4090 GPU.

\section{Discussion}

\noindent\textbf{Limitations.}
Despite the demonstrated effectiveness of Task-Edit, some limitations remain. First, Task-Edit primarily handles rigid objects and struggles with non-rigid objects, such as clothes or towels, as well as articulated objects like drawers. This constraint arises because Task-Edit relies on 6-DoF trajectories reconstructed by FoundationPose \cite{foundationpose}, which is currently incapable of processing these two categories of objects. Furthermore, Task-Edit depends on image-based 3D generation methods \cite{hunyuan3d, rodin} to create 3D models. Although these methods effectively reconstruct the texture and geometry of common objects, their physical scales often deviate from real-world dimensions, necessitating manual adjustment.

\noindent\textbf{Future Works.}
Task generation refers to creating new tasks from existing task demonstrations. For instance, it generates a task like ``pushing a bottle to a specified location'' from an original demonstration of ``placing a beverage box in a specified location.'' In this context, Task-Edit serves as an essential intermediate step toward comprehensive task generation. Addressing intra-task diversity in the current setting is a prerequisite for advancing to task generation scenarios with higher data diversity. Furthermore, integrating non-rigid and articulated objects into Task-Edit, as well as developing more effective methods to mitigate the Sim2Real gap, remains a valuable direction for future research.

\section{Conclusion}
In this paper, we present Task-Edit, a data-efficient framework that generates diverse data from a task-centric editing perspective. By decomposing tasks into scene, skill and object components, our framework enables the generation of diverse, high-quality demonstration sets through flexible combinations. Extensive real-world experiments confirm that Task-Edit improves the performance of 3D visuomotor policies across various embodiments and tasks. Our results also demonstrate that Task-Edit enhances model generalization and strengthens applicability in challenging scenarios that are difficult to capture through manual collection. Above all, Task-Edit provides a scalable and efficient solution to the data scarcity and diversity problem in 3D visuomotor policy learning, particularly for complex and long-horizon tasks.

\bibliography{reference.bib}
\bibliographystyle{IEEEtran}

\vfill

\end{document}